# Guidelines for the Annotation and Visualization of Legal Argumentation Structures in Chinese Judicial Decisions

*Working Draft v1.1(June 2026)*

Kun Chen[1], Xianglei Liao[1], Kaixue Fei[1], Yi Xing[1], Xinrui Li[1]

## Abstract

This Guideline aims to construct a systematic, operational, and reproducible annotation framework for legal argumentation structures, designed to reveal the logical structure of judicial reasoning in court decisions. Grounded in legal argumentation theory, the Guideline establishes formal representation rules and diagrammatic standards through clearly defined categories of proposition types and inter-propositional relation types. It further incorporates rigorous annotation procedures and consistency control mechanisms, thereby providing a reliable data foundation for subsequent automated analysis, structural mining, and model training.

At the level of elements, the Guideline distinguishes between the non-propositional layer and the propositional layer. The non-propositional layer consists of two elements: Issue (IS) and Non-argumentative Components (Non). At the proposition level, the Guideline distinguishes four types: General Normative Judgments (GM), Particular Normative Judgments (SM), General Factual Judgments (GF), and Particular Factual Judgments (SF), and sets out principles for sub-classifying GM. The annotation process emphasizes principles such as *prioritized matching*, *non-duplication*, and *reliance on explicit textual expression* to ensure stability and consistency in type identification.

At the relational level, five relation types are defined: Support Relation (S), Attack Relation (A), Joint Relation (J), Match Relation (M), and Identity Relation (I). These relations capture the logical connections among propositions within judicial reasoning. Support and attack relations represent positive and negative argumentative directions; joint relations characterize conjunctive structures in which multiple propositions are jointly necessary; match relations describe the correspondence between general rules and specific case facts; and identity relations address semantic equivalence. The Guideline further specifies formal expression rules for each relation type, distinguishes between non-nested and multi-level nested structures, and establishes strict hierarchical representation rules.

At the diagrammatic level, the Guideline introduces unified visualization standards: proposition nodes are represented by rectangles, relation nodes by circles, and different circle styles (solid, hollow, and "+") distinguish support, attack, joint, and match relations, respectively; identity is represented by a slash "/" within a rectangular node. For nested structures, diagrams must be constructed layer by layer from the innermost relation outward, preserving the independence and hierarchical clarity of each relational node, thereby ensuring readability and verifiability of structural representations.

[1] Law School, Nanjing University. Email: chenkun@nju.edu.cn.

In addition, the Guideline provides a comprehensive example demonstrating the full annotation workflow—from determining annotation scope and segmenting propositions to type annotation, relation annotation, and argument diagram construction—explicitly illustrating the correspondence between formal representations and diagrammatic methods, thus enhancing practical applicability.

Overall, this Guideline establishes a structurally coherent, logically rigorous, and formally unified framework for annotating legal argumentation. It provides methodological foundations and technical standards for the quantitative analysis, computational modeling, and intelligent applications of judicial reasoning structures.

# 1. Motivation, Objectives, Significance, and Target Users

## 1.1 Motivation

In judicial adjudication, legal reasoning occupies a central position. The core of legal evaluation lies not merely in the conclusion of a judgment, but in whether the reasons supporting that conclusion are sufficient, whether the reasoning process is justified, and whether the argumentation is acceptable. For this reason, judicial practice in China has long emphasized the explanatory function of judicial decisions. In 2018, the Supreme People's Court issued the Guiding Opinions on Strengthening and Regulating the Reasoning in Judicial Documents, systematically clarifying the importance and normative requirements of reasoned adjudication.

With the ongoing advancement of digital governance and the rapid development of judicial artificial intelligence, the reasoning function of judicial documents has been situated within a new technological and institutional context. Judicial decisions are increasingly used in scenarios such as similar-case retrieval, judgment evaluation, legal supervision, and decision support. In these applications, the internal structure of legal argumentation embedded in judicial documents becomes a crucial entry point for understanding judicial decision-making mechanisms, evaluating adjudicative quality, and enabling intelligent judicial systems.

However, most existing research in judicial artificial intelligence has focused primarily on outcome prediction or document generation. The treatment of judicial reasoning typically remains at the level of surface-level textual or statistical features, lacking a fine-grained representation of argumentative structure, inferential relations, and normative constraints. Although such approaches may achieve certain technical performance gains, they struggle to meet the intrinsic demand for explainability in judicial decision-making and are insufficient to support auditable intelligent adjudication systems.

Against this background, systematically annotating the structure of legal argumentation in judicial decisions becomes a necessary bridge between doctrinal legal theory and computational legal studies. By explicitly identifying argumentative elements and their interrelations within judicial texts, the inferential processes that are otherwise implicit in natural language can be transformed into structured objects that are observable, analyzable, and computationally tractable. This transformation not only provides an empirical basis for validating and refining theories of legal argumentation, but also establishes the data and methodological foundations for subsequent research on argument mining, argument evaluation, argumentative conclusion computation, and automated reasoning generation, thereby promoting the development of explainable legal artificial intelligence.

## 1.2 Objectives

This Guideline aims to propose a structured annotation and visualization framework for legal argumentation in the reasoning sections of Chinese judicial decisions. It is designed to provide a unified, reproducible, and extensible operational standard for manual annotation, dataset construction, and related research.

Specifically, the objectives of this Guideline are as follows:

- **To define the fundamental elements of legal argumentation and establish a corresponding attribute labeling scheme.** By distinguishing different types of argumentative components within judicial decisions, this Guideline identifies and annotates their respective properties and functional roles in the argumentative process, thereby providing clear analytical units for understanding the internal composition of judicial reasoning.
- **To specify the types of relationships among argumentative elements and formulate corresponding annotation rules.** The framework focuses on capturing core relational structures such as support, attack, joint, and match, thereby revealing the modes and pathways through which reasoning unfolds in judicial justifications.
- **To represent the overall structure of legal argumentation through graphical representations.** Based on element-level and relation-level annotations, the argumentative process embedded in judicial decisions is transformed into an intuitive and interpretable structured diagram, illustrating the logical trajectory through which adjudicative conclusions are derived.
- **To establish a unified foundation for subsequent research and applications.** By developing a standardized annotation and visualization framework, this Guideline provides the data and methodological basis for future studies on automated argument mining, argument evaluation, argumentative conclusion computation, and argument generation.

## 1.3 Significance

Annotating and visualizing the structure of legal argumentation in judicial decisions holds substantial significance for legal education, legal research, the standardization of judicial reasoning, and the development of legal artificial intelligence.

### 1.3.1 Legal Education

In legal education, judicial decisions have long served as essential materials for cultivating legal reasoning and argumentative competence. However, students often struggle to grasp the internal structure of judicial justifications, tending to focus primarily on final conclusions or isolated argumentative fragments.

By introducing structural annotation and graphical representation of legal argumentation, this Guideline provides an intuitive and operational pedagogical tool for teaching judicial decisions. Through visualized argumentative structures, students can clearly observe the inferential relationships among factual propositions, normative propositions, and adjudicative conclusions, thereby gaining a more comprehensive understanding of how judicial reasoning unfolds.

This approach not only enhances students' ability to read and analyze judicial decisions but also fosters the development of structured and normatively disciplined legal reasoning in their own writing and argumentation.

### 1.3.2 Legal Research

For legal scholars, this Guideline offers a structured, comparable, and accumulative analytical framework for the study of judicial reasoning. Traditionally, research on judicial justification and legal argumentation has relied heavily on abstract doctrinal analysis or case-by-case examination. As a result, conclusions often rest on holistic interpretation or experiential judgment, making systematic comparison across cases difficult.

By standardizing the annotation of argumentative elements and their interrelations, this Guideline enables the decomposition, reconstruction, and reproduction of reasoning processes embedded in judicial texts. It thus provides empirically examinable research materials for theories of legal argumentation, legal methodology, and empirical studies of judicial justification.

Such an approach facilitates the identification of structural similarities and differences across cases, judges, and courts, promoting a shift in legal research from purely abstract analysis toward more fine-grained and empirically grounded investigation.

### 1.3.3 Standardization of Judicial Reasoning

At the level of judicial practice, the annotation and visualization of legal argumentation structures contribute to enhancing the transparency and examinability of judicial reasoning. By presenting justificatory structures in a structured format, the argumentative foundations of adjudicative conclusions can be more clearly displayed.

This structured representation assists litigants and the broader public in understanding judicial logic, thereby providing a rational basis for evaluating adjudicative quality and strengthening judicial supervision.

### 1.3.4 Development of Judicial Artificial Intelligence

In the development of judicial artificial intelligence, the annotation of legal argumentation structures plays a foundational and prerequisite role. Unlike outcome-oriented predictive models, AI systems designed for judicial applications must be capable of articulating their reasoning grounds and argumentative pathways in order to satisfy legal requirements of explainability, normativity, and accountability.

By systematically annotating argumentative elements, relational structures, and overall reasoning configurations, this Guideline establishes a unified data foundation and methodological reference for automated argument mining, representation, computation, evaluation, and generation.

This framework not only supports the training and validation of explainable judicial AI models but also provides explicit criteria for assessing their rationality and compliance, thereby constructing a stable interface between technological development and the normative demands of the rule of law.

## 1.4 Target Users

The annotation and visualization framework for legal argumentation structures proposed in this Guideline is applicable to a variety of research and practical scenarios centered on the reasoning sections of Chinese judicial decisions. It is primarily intended for the following audiences:

- **Legal Scholars.** This Guideline is suitable for researchers engaged in the study of judicial justification, legal argumentation, legal methodology, and judicial institutions.

By employing the proposed framework, scholars can conduct systematic and structured analyses of the reasoning processes embedded in judicial decisions, thereby generating reproducible analytical results in case studies, comparative research, and empirical investigations.

- **Law Teachers and Students.** This Guideline is applicable to instructors and learners who use judicial decisions as teaching materials in legal education, particularly in courses on legal methodology, case analysis, and the close reading of judicial opinions. The annotation and visualization of judicial reasoning structures enable learners to intuitively grasp the unfolding process of legal argumentation, thereby enhancing their ability to read judicial decisions and to construct well-structured legal arguments.
- **Members of the Public and Legal Practitioners.** This Guideline is relevant to individuals engaged in the evaluation of adjudicative quality. Through the structured annotation and graphical representation of legal argumentation in judicial decisions, users can better understand the logical organization of judicial reasoning and obtain standards and reference points for assessing the quality of judicial opinions.
- **Researchers in Law and Artificial Intelligence.** This Guideline is designed for technical researchers and interdisciplinary teams working at the intersection of law and artificial intelligence, particularly those engaged in tasks such as automated argument mining, evaluation, computation, and generation. The framework may serve as a foundational specification for manual annotation, dataset construction, and model evaluation, enabling such research to meet technical objectives while maintaining sensitivity to legal normativity and explainability requirements.

# 2. Annotation Targets, Scope, and Granularity

## 2.1 Annotation Targets

The primary annotation targets of this Guideline are judicial decision documents, including:

- Original judgments published on China Judgments Online (https://wenshu.court.gov.cn);
- Guiding cases and reference cases published in the People's Court Case Database (https://rmfyalk.court.gov.cn/).

These documents constitute the textual corpus within which legal argumentation structures are to be identified and annotated.

## 2.2 Annotation Scope

This Guideline distinguishes between *legal argumentation annotation* and *case information annotation*, each serving distinct analytical purposes.

### 2.2.1 Scope of Legal Argumentation Annotation

The scope of legal argumentation annotation is limited to the reasoning sections of original judgments, guiding cases, and reference cases.

To further clarify the boundaries of annotation, examples drawn from original judgments, guiding cases, and reference cases will be provided below. These examples serve to illustrate the specific textual segments that fall within the annotation scope and to delineate the portions of judicial documents that are excluded from structural argumentation annotation.

**Document I: Example of an Original Judgment**

Civil Judgment (First Instance) in a Labor Contract Dispute between Yu and Xu

People's Court of Gulou District, Nanjing, Jiangsu Province Civil Judgment

(2023) Su 0106 Min Chu No. 18909

Plaintiff: Yu.

Defendant: Xu.

The plaintiff Yu filed a lawsuit against the defendant Xu in a labor contract dispute. After the case was docketed on December 8, 2023, this Court applied the small-claims procedure in accordance with the law and conducted a public hearing. The plaintiff Yu appeared in court to participate in the proceedings. The defendant Xu, having been duly summoned by this Court, failed to appear in court without justified reason. The case has now been concluded.

The plaintiff Yu submitted the following claims to this Court: the defendant shall pay the plaintiff labor remuneration in the amount of RMB 11,600.

Facts and reasons: From 2019 to the end of 2021, the plaintiff was employed by the defendant to perform work in Gulou District, Nanjing, during which part of the labor remuneration was paid intermittently. After the completion of the project, the parties conducted a settlement and confirmed that RMB 17,000 remained unpaid as the outstanding balance of labor remuneration. On January 10, 2022, the defendant signed an IOU stating: "I hereby owe the plaintiff labor wages in the amount of RMB 17,000." On January 17, 2022, the plaintiff was again employed by the defendant to carry materials at Building No. 6 in Nanjing, with labor remuneration totaling RMB 600. At this point, the defendant owed the plaintiff a total of RMB 17,600 in labor remuneration. On January 30, 2022, the defendant transferred RMB 6,000 to the plaintiff via WeChat as payment of labor remuneration. Up to the present date, the defendant still owes the plaintiff RMB 11,600 in labor remuneration. The plaintiff has repeatedly demanded payment, but the defendant has refused to answer phone calls or reply to WeChat messages in an attempt to evade responsibility. In order to safeguard the plaintiff's lawful rights and interests, the plaintiff has filed this lawsuit and requests that the Court render judgment as prayed.

The defendant Xu did not appear in court.

Upon trial, this Court ascertains the following facts: On January 10, 2022, the defendant issued an IOU to the plaintiff, stating: "I owe Yu labor wages in the amount of RMB 17,000."

On January 30, 2022, a third party, Xu (whom the plaintiff stated to be the son of Xu), transferred RMB 6,000 to the plaintiff via WeChat.

During the hearing, the plaintiff stated that the additional RMB 600 was remuneration for services provided after the issuance of the IOU. The corresponding evidence included a location sent by the defendant to the plaintiff via WeChat, which indicated the address where the services were performed.

This Court holds that <u>a legally established contract shall take effect upon its formation. The parties shall fully perform their respective obligations in accordance with the agreement. Where one party fails to pay the agreed price, remuneration, rent, interest, or fails to perform other monetary obligations, the other party may request payment. In this case, a labor contract relationship existed between the plaintiff and the defendant. The</u>

plaintiff has performed labor in accordance with the agreement, and the defendant shall pay labor remuneration in accordance with the agreement between the parties. Based on the IOU submitted by the plaintiff and the WeChat transfer records, it can be determined that the defendant still owes the plaintiff RMB 11,000 in labor remuneration, and the defendant shall pay the RMB 11,000 to the plaintiff. The plaintiff further claims an additional RMB 600 in labor remuneration; however, the evidence provided is insufficient to substantiate this claim. Therefore, this Court does not support the plaintiff's request that the defendant pay the additional RMB 600 in labor remuneration.

In summary, in accordance with Articles 502, 509, and 579 of the Civil Code of the People's Republic of China, and Article 40(2), Article 67(1), Article 147, and Article 165 of the Civil Procedure Law of the People's Republic of China, the judgment is as follows:

The defendant Xu shall pay the plaintiff Yu labor remuneration in the amount of RMB 11,000 within three days from the effective date of this judgment;

The plaintiff Yu's other claims are dismissed.

If the monetary obligation is not performed within the period specified in this judgment, interest on the debt during the period of delayed performance shall be doubled in accordance with Article 260 of the Civil Procedure Law of the People's Republic of China.

The case acceptance fee of RMB 45 shall be borne as follows: RMB 7.50 by the plaintiff Yu and RMB 37.50 by the defendant Xu.

This judgment is final.

Judge: XX

December 28, 2023

Clerk: XXX

From the above example of the judgment, it can be observed that original judgments do not contain explicit structural labels. However, judgments drafted in accordance with the document format issued by the Supreme People's Court can generally be divided into three main parts: the introductory part (including the title, case number, party information, and case overview), the main body (including facts, reasoning, and the decision), and the concluding part (including the allocation of litigation costs, signatures, and date, etc.).

The court's reasoning is reflected in the portion of the main body that appears after the phrase "This Court holds that" and before the sentence introducing the dispositive section, namely "The judgment is as follows."

**Document II: Example of a Guiding Case**

Shanghai Centaline Property Consultants Co., Ltd. v. Tao Dehua — Dispute over an Intermediary Contract

(Discussed and adopted by the Judicial Committee of the Supreme People's Court and released on December 20, 2011)

**Keywords:** Civil; Intermediary Contract; Second-hand Housing Sale; Breach of Contract

**Adjudication Highlights**

In a housing sale intermediary contract, a clause prohibiting the buyer from using housing information provided by the intermediary company while bypassing that intermediary to conclude a housing sale contract with the seller is lawful and valid. However, where the seller lists the same property for sale through multiple intermediary companies, and the buyer obtains the same housing information through other legitimate and publicly accessible channels, the buyer has the right to choose an intermediary company offering a lower quotation and better services to facilitate the conclusion of the housing sale contract. In such circumstances, the buyer's conduct does not constitute the use of housing information provided by the previously contracted intermediary company and therefore does not constitute a breach of contract.

**Relevant Legal Provision**

Article 424 of the Contract Law of the People's Republic of China

**Basic Facts of the Case**

The plaintiff, Shanghai Centaline Property Consultants Co., Ltd. (hereinafter referred to as "Centaline"), alleged that: the defendant Tao Dehua used the housing sale information for a property located at No. ___, Zhuzhou Road, Hongkou District, Shanghai, which had been provided by Centaline, deliberately bypassed the intermediary, and privately concluded a housing purchase contract directly with the seller, thereby violating the agreement in the "Real Estate Purchase Confirmation Letter." This constituted malicious "order-jumping" behavior. Centaline requested that the court order Tao Dehua to pay liquidated damages of RMB 16,500 in accordance with the agreement.

The defendant Tao Dehua argued that: the original property owner, Li, had entrusted multiple intermediary companies to sell the property involved in the case. Centaline did not exclusively possess the housing information, nor was it the exclusive sales agent. Tao Dehua did not use information provided by Centaline, and no "order-jumping" breach of contract occurred.

Upon trial, the court ascertained the following facts: In the second half of 2008, the original property owner Li listed the property for sale with multiple real estate intermediary companies. On October 22, 2008, a Shanghai real estate brokerage company showed the property to Tao Dehua; on November 23, a Shanghai real estate consultancy company (hereinafter referred to as "the consultancy company") showed the property to Tao Dehua's wife, Cao; on November 27, Centaline showed the property to Tao Dehua and, on the same day, signed a "Real Estate Purchase Confirmation Letter" with him.

Article 2.4 of the Confirmation Letter provided that within six months after inspecting the property, if Tao Dehua, or persons associated with him (including entrusted persons, agents, representatives, or other related individuals), used information or opportunities provided by Centaline but concluded a transaction with a third party without going through Centaline, Tao Dehua shall pay Centaline liquidated damages equal to 1% of the actual transaction price agreed upon with the seller.

At that time, Centaline quoted a price of RMB 1.65 million for the property, whereas the consultancy company quoted RMB 1.45 million and actively negotiated the price with the seller. On November 30, under the mediation of the consultancy company, Tao Dehua and the seller signed a housing sale contract at a transaction price of RMB 1.38 million. Thereafter, the parties completed the transfer procedures, and Tao Dehua paid the consultancy company a commission of RMB 13,800.

**Judgment Results**

On June 23, 2009, the Hongkou District People's Court of Shanghai rendered Civil Judgment (2009) Hong Min San (Min) Chu No. 912, ordering that the defendant Tao Dehua shall, within ten days from the effective date of the judgment, pay the plaintiff Centaline liquidated damages of RMB 13,800.

After the pronouncement of the judgment, Tao Dehua filed an appeal. On September 4, 2009, the Shanghai No. 2 Intermediate People's Court rendered Civil Judgment (2009) Hu Er Zhong Min Er (Min) Zhong No. 1508, ruling as follows:

The Civil Judgment (2009) Hong Min San (Min) Chu No. 912 of the Hongkou District People's Court of Shanghai is revoked;

Centaline's claim requesting Tao Dehua to pay liquidated damages of RMB 16,500 is not supported.

**Court's Reasoning**

The effective judgment held that the "Real Estate Purchase Confirmation Letter" signed between Centaline and Tao Dehua was in the nature of an intermediary contract. Article 2.4 thereof constituted a commonly used standard clause in housing sale intermediary contracts prohibiting "order-jumping." Its purpose was to prevent a buyer from using housing information provided by the intermediary company while bypassing the intermediary to purchase the property, thereby depriving the intermediary of its due commission. This clause did not exempt one party from liability, increase the other party's liability, or exclude the other party's principal rights, and should therefore be deemed valid.

According to this clause, the key to determining whether the buyer committed "order-jumping" in breach of contract lies in whether the buyer used the housing information or opportunities provided by the intermediary company. If the buyer did not use the information or opportunities provided by that intermediary company but instead obtained the same housing information through other legitimate and publicly accessible channels, the buyer has the right to choose an intermediary offering a lower quotation and better services to facilitate the conclusion of the housing sale contract, and such conduct does not constitute "order-jumping" in breach of contract.

In this case, the original property owner listed the same property for sale through multiple intermediary companies. Tao Dehua and his family obtained information about the same property through different intermediary companies and ultimately concluded the housing sale contract through another intermediary. Therefore, Tao Dehua did not use the information or opportunities provided by Centaline and did not constitute a breach of contract. Accordingly, Centaline's claims were not supported.

**Document III: Example of a Reference Case**

Huayin City XX Shareholding Economic Cooperative v. Huayin City XX Comprehensive Market Management Service Center — Dispute over a Lease Contract

— The portion of the lease term of collectively owned construction land exceeding twenty years is invalid

**Keywords:** Civil; Lease Contract; Collectively Owned Construction Land; Lease Term;

Twenty Years

**Basic Facts**

The plaintiff, Huayin City XX Shareholding Economic Cooperative (hereinafter referred to as "the Cooperative"), alleged that: the "Agreement on Leasing Land for the Construction of the South Ring XX Market" and the "Supplementary Agreement on Leasing Land for the South Ring XX Market" signed between the Cooperative and Huayin City XX Comprehensive Market Management Service Center (hereinafter referred to as "the Service Center") both stipulated a lease term of thirty years. Pursuant to Article 705 of the Civil Code of the People's Republic of China, which provides that "where the lease term exceeds twenty years, the portion exceeding twenty years shall be invalid," the agreed lease term exceeding twenty years should be deemed invalid, and the Service Center should return the leased land. The Cooperative therefore filed suit, requesting that the court: (1) declare invalid the portion of the lease term exceeding twenty years in the agreements signed on March 8, 2001, namely the period from July 1, 2021 to June 30, 2030; and (2) terminate the contracts at issue and order the payment of rent, etc.

The defendant Service Center argued that: the subject matter of the contracts at issue was land and was not subject to the twenty-year limitation. According to the contract, the Service Center leased land from Nansi Village to construct buildings. Article 4 of the Ministry of Land and Resources' "Opinions on Regulating the Leasing of State-Owned Land" (Guo Tu Zi Fa [1999] No. 222) should apply by analogy. The Service Center's construction of buildings after leasing the land should be treated by reference to renewal of a long-term lease. The agreed thirty-year land use term did not exceed the maximum term for the transfer of land use rights for similar purposes and complied with the relevant requirements.

Upon trial, the court found that on March 8, 2001, the Nansi Group of XX Village, Huayin City (Party A), and the Service Center (Party B) signed the "Agreement on Leasing Land for the Construction of the South Ring XX Market," which provided: (1) In order to promote Party A's commodity market, Party B's preparatory office shall lease 1.79 mu of land from Party A; … (2) Party B's preparatory office shall construct 16 vegetable greenhouses, a road 9 meters wide and 78 meters long running north to south, sidewalks, and other works; (3) The annual rent shall be RMB 1,500 per mu, totaling RMB 2,700 per year; (4) Payment shall be made in full by the end of December each year; (5) The lease term shall be thirty years, from July 1, 2001 to the end of June 2030. The agreement bore the seals of the XX Village Committee of Huayin City and the Service Center, as well as the signatures and seals of relevant individuals.

On May 18, 2001, the same parties signed the "Supplementary Agreement on Leasing Land for the South Ring XX Market," which provided in Article 4 that the lease term shall be thirty years, from July 1, 2001 to June 30, 2030. The agreement bore the corresponding seals and signatures.

On May 14, 2018, the Service Center paid RMB 23,850 for the land of the South Ring XX Market, covering the period from May 2018 to May 2020. A receipt for payment by a rural collective economic organization of Huayin City was issued.

It was further ascertained that the Nansi Group of XX Village is the same as Group Three

of XX Village, Huayin City, and that the Cooperative is the collective economic organization established by Group Three of that village. The current land use map of Huayin City shows that the land involved is classified as construction land.

On June 7, 2023, the People's Court of Huayin City, Shaanxi Province rendered Civil Judgment (2023) Shaan 0582 Min Chu No. 128, ruling: (1) the portion of the lease term exceeding twenty years in the agreements signed on March 8, 2001 and May 18, 2001, namely the period from July 1, 2021 to June 30, 2030, is invalid; (2) the contracts at issue are terminated, the Service Center shall return all the leased land and pay rent of RMB 21,389 for the period from June 1, 2020 to February 7, 2023, as well as a land occupation fee of RMB 21.78 per day from February 8, 2023 until the actual return of the land; and (3) the Cooperative's other claims are dismissed.

After the judgment was pronounced, the Service Center filed an appeal. On November 28, 2023, the Intermediate People's Court of Weinan City, Shaanxi Province rendered Civil Judgment (2023) Shaan 05 Min Zhong No. 2039, dismissing the appeal and upholding the original judgment.

**Court's Reasoning**

The disputed issues in this case are: (1) whether the portion of the lease term of the collectively owned construction land exceeding twenty years is valid; and (2) when the contracts at issue were terminated.

Regarding the first issue, Article 361 of the Civil Code of the People's Republic of China provides that collectively owned land used for construction shall be handled in accordance with land administration laws. Article 63(4) of the Land Administration Law of the People's Republic of China provides that the leasing of collectively owned commercial construction land, as well as the grant, maximum term, transfer, exchange, capital contribution, gift, and mortgage of the right to use collectively owned construction land, shall be implemented by reference to state-owned construction land of similar use, with specific measures to be formulated by the State Council. Article 467(1) of the Civil Code provides that where a contract is not expressly provided for in this Code or other laws, the general provisions of this Book shall apply, and the provisions on the most similar type of contract may be applied by reference. Article 705 provides that the lease term shall not exceed twenty years, and any portion exceeding twenty years shall be invalid.

The land involved in the agreements at issue is construction land. The contracts signed by the parties constitute lease contracts for collectively owned construction land as provided by law. The lease term should be subject to measures formulated by the State Council. As the State Council has not yet formulated relevant measures or provisions, the lease term of collectively owned construction land shall be handled in accordance with the provisions of the Civil Code concerning lease terms. Accordingly, the portion of the lease term exceeding twenty years in the agreements at issue is invalid.

Regarding the second issue, Article 734 of the Civil Code provides that where the lease term expires and the lessee continues to use the leased property without objection from the lessor, the original lease contract remains valid but the lease term becomes indefinite. After the agreements at issue expired on June 30, 2021, the Service Center continued to occupy and use the market land, and the Cooperative did not take measures or raise objections. Therefore, the contracts continued to be valid, and the lease term became indefinite.

Article 730 of the Civil Code provides that where the parties have not agreed on the lease term or the agreement is unclear and cannot be determined pursuant to Article 510, the lease shall be deemed indefinite; either party may terminate the contract at any time upon reasonable notice. Article 565 provides that where a party claims termination of a contract by filing a lawsuit or applying for arbitration without notifying the other party, and the people's court or arbitral institution confirms the claim, the contract shall be terminated upon service of the copy of the complaint or arbitration application on the other party.

In this case, during the performance of the indefinite lease contract, the Cooperative filed a lawsuit in February 2023 requesting termination of the contract. The Service Center received a copy of the complaint on February 7, 2023. Therefore, in accordance with the above provisions, the lease contract between the parties shall be deemed terminated on February 7, 2023. In light of the facts concerning land occupation in this case, the court rendered the above judgment according to law.

**Adjudication Summary**

According to the Land Administration Law of the People's Republic of China, the lease term of collectively owned construction land shall be implemented by reference to state-owned construction land of similar use, with specific measures to be formulated by the State Council. As the State Council has not formulated relevant measures or provisions concerning the lease term of state-owned construction land, the lease term of collectively owned construction land shall be governed by the provisions of the Civil Code concerning lease terms. Accordingly, the lease term of collectively owned construction land shall not exceed twenty years, and any portion exceeding twenty years shall be invalid.

**Related Provisions**

Civil Code of the People's Republic of China: Articles 361, 467, 705

Land Administration Law of the People's Republic of China (2019 Amendment): Article 63

First Instance: Civil Judgment (2023) Shaan 0582 Min Chu No. 128, People's Court of Huayin City, Shaanxi Province (June 7, 2023)

Second Instance: Civil Judgment (2023) Shaan 05 Min Zhong No. 2039, Intermediate People's Court of Weinan City, Shaanxi Province (November 28, 2023)

It can be observed that, in guiding cases and reference cases, the reasoning section has clear textual boundaries.

In Document I, Document II, and Document III above, the underlined portions indicate the scope of legal argumentation annotation.

### 2.2.2 Case Information Annotation

To facilitate retrieval, organization, and statistical analysis, the following case information is annotated for original judgment documents: case number, adjudicating court, court level, trial level, date of judgment, case category, cause of action, and type of judgment outcome.

**Table 1. Annotated Case Information in Original Judgment Documents**

| Field | Description | Example |
|---|---|---|
| Case Number | The case number indicated at the beginning of the judgment document | (2023) Su 0106 Min Chu No. 18909 |
| Adjudicating Court | Full name of the People's Court rendering the judgment | Nanjing Gulou District People's Court of Jiangsu Province |
| Court Level | Basic / Intermediate / Higher / Supreme | Basic People's Court |
| Trial Level | First instance / Second instance / Retrial | First instance |
| Date of Judgment | Date on which the judgment was rendered | 28-Dec-23 |
| Case Category | Civil / Criminal / Administrative / Enforcement | Civil |
| Cause of Action | General description of the legal relationship involved in the case | Labor contract dispute |
| Type of Judgment Outcome | Fully upheld / Partially upheld / Dismissed | Partially upheld |

For guiding cases, the following case information is annotated: case type, case name, date of release, case category, relevant legal provisions, and adjudication highlights.

**Table 2. Annotated Information for Guiding Cases**

| Field | Description | Example |
|---|---|---|
| Case Type | Guiding Case / Reference Case | Guiding Case |
| Case Name | Title of the case | Shanghai Centaline Property Consultants Co., Ltd. v. Tao Dehua — Dispute over an Intermediary Contract |
| Date of Release | Date on which the case was issued | 20-Dec-11 |
| Case Category | Civil / Criminal / Administrative / Enforcement | Civil |
| Relevant Legal Provisions | Relevant legal provisions expressly cited in the case document | Article 424 of the Contract Law of the People's Republic of China |
| Adjudication Highlights | The principal adjudicative rule of the case | Where a seller lists the same property for sale through multiple intermediary companies, and the buyer obtains the same housing information through other legitimate and publicly accessible channels, the buyer has the right to choose an intermediary offering a lower quotation and better services to conclude the housing sale contract. Such conduct does not constitute the use of housing information provided by the previously contracted intermediary company and therefore does not constitute a breach of contract. |

For reference cases, the following case information is annotated: case type, case name, database entry number, case category, relevant legal provisions, and adjudication highlights.

**Table 3. Annotated Information for Reference Cases**

| Field | Description | Example |
|---|---|---|
| Case Type | Guiding Case / Reference Case | Reference Case |
| Case Name | Title of the case | Huayin City XX Shareholding Economic Cooperative v. Huayin City XX Comprehensive Market Management Service Center — Dispute over a Lease Contract |
| Database Entry Number | The reference case number in the People's Court Case Database | 2025-07-2-111-002 |
| Case Category | Civil / Criminal / Administrative / Enforcement | Civil |
| Relevant Legal Provisions | Legal provisions cited in the associated index section of the judgment | Articles 361, 467, and 705 of the Civil Code of the People's Republic of China; Article 63 of the Land Administration Law of the People's Republic of China (2019 Revision) |
| Adjudication Highlights | The adjudicative rule of the case | The lease term for collectively owned construction land shall be governed by the relevant provisions on lease terms under the Civil Code. |

## 2.3 Granularity

For the legal argumentation annotation regulated by this Guideline, the initial segmentation granularity is uniformly defined at the level of sentences separated by periods (。) or semicolons (;). After this first-stage segmentation, each clause separated by commas (,) is examined individually (in some cases, a complete proposition may only be expressed through multiple clauses).

If consecutive clauses are of the same type and there is no explicit support or opposition relation among them, these clauses are annotated as a single element and assigned one element identifier. By contrast, if consecutive clauses differ in type or exhibit a clear Support or Attack relation, they cannot be treated as a single element; instead, they must be annotated separately, with each assigned its own element identifier. Importantly, in the annotation of propositional elements, a proposition refers to a truth-evaluable statement. In judicial texts, however, judges do not always express propositions in declarative form, and annotators are therefore required to extract the underlying proposition. For example:

> *From Article 2 of the Supplementary Agreement signed by both parties, which stipulates that "the exact area specified in the lease contract dated June 9, 2007 shall be determined according to the actual area in use, including both the area covered by property title certificates and the area without such certificates"*

For the basic case information, the basic annotation unit may, as needed, be a word, a sentence, or a paragraph.

# 3. Overall Annotation Approach

This Guideline adopts a four-stage annotation procedure involving granularity alignment, element annotation, inter-propositional relation annotation, and argument graph generation. Adjudicators may intervene at any stage during the annotation process of a case. The annotation of a case proceeds sequentially through four stages—granularity alignment, element annotation, relation annotation, and graph generation—with historical versions of each case annotation retained at every stage for subsequent data analysis. In addition, propositional roles (such as premise, sub-conclusion, and conclusion) are not treated as independent annotation items; rather, they are automatically inferred through graph analysis based on the relational structure among propositions.

**Granularity Alignment:** Annotators independently segment the reasoning section of the judgment in accordance with the current granularity alignment scheme. Upon completion, the annotated case document is represented specifically in the form of boxed text spans within the original text area. At this stage, neither elements nor relations are annotated. The case document is then submitted to the arbitrator. After obtaining the segmentation documents produced by the annotators, the arbitrator compares them: if the discrepancy between the two annotators is within 10%, the process proceeds to the next stage; if it exceeds 10%, the annotators are required to perform segmentation again.

**Element Annotation:** Annotators perform element annotation on the segmented text by assigning each clause an element type and a serial number, thereby generating an element-area Word table, which is submitted to the arbitrator for review. The arbitrator examines element consistency. If disagreement remains below 10%, the process proceeds to relation annotation; if inconsistency exceeds 10%, an online meeting is convened for discussion.

**Relation Annotation:** Based on the finalized elements, annotators conduct relation annotation sequentially according to the order of elements. The annotation results form the relation area, which is then submitted to the arbitrator for review. Consistency is controlled within 15%. If inconsistency in annotation results exceeds 15%, an online meeting is required for deliberation.

**Graph Generation:** Corresponding argument graphs are generated based on each case's relation area, and the final argument graph results are submitted to the arbitrator for review. Once it is confirmed that there are no misrepresentations, symbol errors, or omissions, the argument graph is considered complete.

# 4. Element Type System

## 4.1 Non-propositional Element Types

Non-propositional elements comprise two categories: Issue (IS) and Non-argumentative Elements (Non). IS is used to annotate section headings, as well as the court's summarizing statements regarding the disputes in a case. Non is used to annotate information that is unrelated to the argumentation.For example:

*"The disputed issues in this case are threefold: whether the alleged coverage constituted complete coverage or partial coverage; if partial coverage, how the scope of coverage should be determined; how the scope, standards, amount, and interest of coverage compensation should be determined; and whether a certain railway bureau should bear joint liability with a certain railway company for payment of the compensation at issue."*

These are annotated as IS. Information unrelated to the present argumentation is annotated as Non.

## 4.2 Proposition Type System

### 4.2.1 Design Principles

The construction of the propositional type system should serve the stable identification and analysis of the argumentative structure of judicial decisions. In the design process, this Guideline adheres to the following principles.

**Principle of Minimal Distinction:** The classification of proposition types shall be limited to what is necessary for argumentative structure analysis, distinguishing only those categories of propositions that are decisive or significantly influential for the development of reasoning. Propositions with minor semantic differences and no substantive impact on the reasoning structure shall not be excessively subdivided, so as to avoid forming an overly complex and impractical type system.

**Principle of Grounding in Judicial Document Expressions:** The proposition type system is not a reconstruction of legal theoretical structures, but a formalized organization of the actual modes of reasoning employed in judicial documents. Therefore, the division and identification of sub-types are based on the expressions actually used in judicial documents, rather than on any presupposed jurisprudential theoretical model.

**Supporting Subsequent Automated Processing:** The proposition type system shall provide a stable intermediate representation for subsequent tasks such as argument mining, structural computation, and text generation. Accordingly, the classification of types shall be identifiable and reproducible; the standards for classification shall be as formalized as possible; the number of types shall be controlled within a computationally manageable range; and the system shall not rely on highly subjective interpretation.

These principles and requirements provide guidance for the future revision or expansion of basic types or sub-types.

### 4.2.2 Two Sets of Basic Distinction Dimensions and Four Basic Types of Judgment

This Guideline first adopts two mutually independent dimensions: "particular" and "general," and "fact" and "norm." Based on these two dimensions, judgments in legal argumentation are divided into four basic types: Particular Factual Judgment (SF), General Factual Judgment (GF), Particular Normative Judgment (SM), and General Normative Judgment (GM) (see Table 4).

**Table 4. Four Basic Types of Judgment**

| | **Particular** | **General** |
|---|---|---|
| **Fact** | Particular Factual Judgment (SF) | General Factual Judgment (GF) |
| **Norm** | Particular Normative Judgment (SM) | General Normative Judgment (GM) |

Functional Positioning of the Four Basic Types of Judgment:

- GF: Provides empirical or background support for the application of legal norms (excluding content from the legal domain) and possesses repeatable applicability across analogous cases (i.e., generality). Examples include social common sense, empirical rules, industry knowledge, and scientific principles.

- GM: Constitutes the normative foundation of legal argumentation and serves as the core element linking case-specific facts with the adjudicative conclusion, including statutory provisions, legal interpretation, and other general normative materials.
- SF: Refers to factual judgments concerning particular objects in the case and serves as the premise for normative application.
- SM: Refers to evaluative judgments with legal significance made by specific relevant entities (including, but not limited to, courts). "Legal significance" refers to judgments concerning the satisfaction of constituent elements and legal consequences in substantive law; procedural determinations and evidentiary determinations in procedural law; as well as judgments regarding illegality, legality, liability, and absence of liability. A particularly distinctive type of Particular Normative Judgment is a contract concluded between the parties.

### 4.2.3 Sub-types of General Normative Judgments (GM)

Among the four basic types of propositions, General Normative Judgments (GM) play the most central role in legal argumentation. To enhance labeling precision and analytical value, GM can be further divided—based on their sources and authority—into five sub-types: statutory provisions, legal interpretation, customs and industry practices, morality and value principles, and other normative judgments (see Table 5).

**Table 5. Five Sub-types of General Normative Judgments**

| Code | Name | Definition | Indicative Expressions |
|---|---|---|---|
| **GM-L** | Statutory Provisions | A broad conception of law is adopted, under which all authoritative normative documents are regarded as legal provisions. This includes both verbatim quotations of authoritative normative texts and restatements of their contents without quotation marks. | "According to Article X of the ___ Law…" "Pursuant to legal provisions…" "The law clearly stipulates that…" |
| **GM-I** | Legal Interpretation | Interpretive judgments concerning the meaning, scope of application, or conditions of application of statutory provisions. Judicial interpretations are classified as statutory provisions rather than legal interpretations. In identifying statutory provisions, the cited content must remain fully consistent with the statutory text; once additional understanding, interpretation, or elaboration is introduced, it shall be classified as legal interpretation. | "It should be understood as…" "It may be interpreted as…" "The intent of this provision is…" |
| **GM-U** | Customs and Industry Practices | Normative judgments derived from social customs, trade practices, or generally accepted industry rules. | "In accordance with trade custom…" "Industry practice is…" "According to industry convention…" |
| **GM-M** | Morality and Value Principles | Normative judgments based on value assessments, public order and good morals, or fundamental principles. | "Violates public order and good morals…" "In consideration of the principle of fairness…" "Contrary to the principle of good faith…" |
| **GM-O** | Other Normative Judgments | General normative judgments that cannot be clearly categorized into the above types. | "According to relevant policy spirit…" "In accordance with XXX administrative regulations…" |

### 4.2.4 Sub-types of Particular Normative Judgments

Among Particular Normative Judgments, contracts possess a distinctive status. Accordingly, contracts are separated from ordinary Particular Normative Judgments (SM) and treated as a special subtype of Particular Normative Judgments, namely Contracts and Contract Interpretation (SM-C).

**Table 6. Two Types of Particular Normative Judgments**

| Code | Name | Definition | Indicative Expressions |
|---|---|---|---|
| SM | Particular Normative Judgments | An evaluative judgment with legal significance made by specific relevant entities (including, but not limited to, courts). | "The plaintiff requests payment of liquidated damages" "The court of first instance held that …" " …lacking factual and legal basis" |
| SM-C | Contracts and Contract Interpretation | Normative content derived from contractual provisions or from normative interpretation of contractual provisions. | "The parties agreed in the contract that …" "According to the contractual agreement …" "This clause should be interpreted as …" |

### 4.2.5 Explanatory Notes

To enhance annotation consistency, the following principles shall be observed when labeling sub-types:

- Priority Matching Principle: If a proposition can be clearly classified under GM-L through GM-M, GM-O shall not be used.
- Non-Redundancy Principle: Each normative judgment shall be assigned only one sub-type.
- Textual Explicitness Principle: Classification shall be based on the explicit textual expression in the document, and the source of the norm shall not be supplemented on the basis of background knowledge or inference.

It should further be clarified that this sub-type system is not a hierarchy of normative authority. Its primary purposes are: (1) to analyze the structural distribution of normative sources in judicial reasoning; and (2) to support subsequent tasks such as automated mining, evaluation, and text generation.

# 5. Types of Inter-Propositional Relations

This Guideline currently specifies five types of relations: support, attack, joint, match, and identity.

The establishment and identification of these relation types aim to reveal the logical connections asserted in judicial reasoning, rather than to reconstruct the actual reasoning process or to evaluate the correctness of the adjudicative conclusion.

## 5.1 Explanation of the Five Relation Types

### 5.1.1 Support Relation

A support relation exists where a single proposition or a group of propositions provides reasons for the establishment of another proposition. In judicial documents, common expressions indicating a support relation include: "therefore," "thus," "accordingly it can be determined that,"

"in summary," and "sufficient to prove," among others.

When a group of propositions supports another proposition, the propositions within the group may either independently support the latter or jointly support it. In the first case, deleting any one proposition from the group does not affect the remaining propositions' support for the target proposition. In the second case, deleting any one proposition would cause the remaining propositions to lose their supporting force. In this situation, the propositions in the group constitute the joint relation described below.

### 5.1.2 Joint Relation

In practice, a single proposition is often insufficient to support a conclusion. Multiple propositions must jointly hold in order to form a complete justification.

This structure of jointly necessary support constitutes a joint relation. The function of the joint relation is to characterize the conjunctive structure formed when multiple propositions support the same conclusion.

### 5.1.3 Match Relation

The match relation is a special case of the joint relation, namely the matching of particular normative/factual judgments with general normative/factual judgments. When the elements of a legal norm correspond to the relevant facts of a case, a matching structure is formed. The match relation typically serves the formation of a particular normative judgment.

The difference between a match relation and an ordinary joint relation lies in the types of propositions involved: ordinary joint relations usually occur among propositions of the same type, and match relations occur between two different types of propositions—namely, general judgments and particular judgments.

### 5.1.4 Attack Relation

An attack relation can only be established after identifying the target proposition or support relation under attack. Logically, an attack relation indicates that the attacked claim or inferential link cannot be accepted on the basis of the available reasons, rather than necessarily asserting the direct negation of the proposition itself.

### 5.1.5 Identity Relation

An identity relation exists where multiple propositions semantically express the same judgment content. This relation is primarily used to address repetitive expressions in judicial documents.

## 5.2 Basic Structures of Different Relation Types — No Nesting

Since the constituent elements of relations such as support and joint may consist of a group of propositions, different types of relations may be nested. The table below presents the basic structures and formal expressions of the five relation types in cases where no nesting occurs (Table 7).

**Table 7. Basic Structures and Formal Expressions of the Five Relation Types**

| Relation Type | Basic Structure | Formal Expression |
|---|---|---|
| **Support (S)** | An ordered pair between any two propositions *pi*, *pj* | *pi* supports *pj*<br>S(pi, pj) |
| **Joint (J)** | A set of propositions {p1, p2, …, pn} (n ≥ 2) | p1, p2, …, pn joint<br>J(p1, p2, ..., pn) |
| **Match (M)** | An ordered pair between a particular judgment *ps* and a general judgment *pg* | *ps* matches *pg*<br>M(ps, pg) |
| **Attack (A)** | An ordered pair between any two propositions *po*, *pc* | *po* attacks *pc*<br>A(po, pc) |
| **Identity (I)** | A set of propositions {p1, p2, …, pn} (n ≥ 2), whose members are semantically equivalent | I(p1, p2, ..., pn) |

## 5.3 Formal Expressions of Different Relation Types — With Nesting

### 5.3.1 Joint Relation Nested within a Match Relation

When a general judgment *pg* contains multiple constituent elements, multiple particular judgments *ps1, ps2, ..., psn* may correspond to pg.

In such a case, *ps1, ps2, ..., psn* form a joint relation among themselves, which is then embedded as a whole within a match relation.

**Representation:** M(J(ps1, ps2, ..., psn), pg)

### 5.3.2 Match Relation Nested within a Joint Relation

The basic form of a joint relation consists of a conjunction among *n* propositions. However, in certain situations, the combined objects are not original propositions but rather several match relations *M1, M2, ..., Mn*.

**Representation:** J(M1, M2, ..., Mn), or J(M(ps1, pg1), M(ps2, pg2), ..., M(psn, pgn))

### 5.3.3 Joint Relation Nested within a Support Relation

The basic form of a support relation is that one proposition provides reasons for another. In practice, however, it frequently occurs that multiple propositions *pi1, pi2, ..., pin* jointly support another proposition *pj*.

In such a case, *pi1, pi2, ..., pin* form a joint relation, which is embedded as a whole within the support relation.

**Representation:** S(J(pi1, pi2, ... pin), pj) (This structure indicates that only when *pi1, pi2, ..., pin* all hold simultaneously can they support *pj*.)

It should be noted that when *n* propositions *pi1, pi2, ..., pin* each independently support *pj*, this constitutes *n* separate support relations rather than a joint relation.

**Representation:** S(pi1, pj)∧S(pi2, pj)∧…S(pin, pj)

The two structures are logically distinct and must not be conflated.

### 5.3.4 Match Relation Nested within a Support Relation

Generally, when a rule applies to a particular case, a structure is formed in which a general judgment *pg* and a particular judgment *ps* (or the conjunction of several particular judgments *ps1, ps2, ..., psn*) support a conclusion *pj*.
In this situation, a match relation is formed between *pg* and *ps* (or the conjunction of *ps1, ps2, ..., psn*), and this match relation is embedded as a whole within a support relation.
**Representation:** S(M(ps, pg), pj), or S(M(J(ps1, ps2, ..., psn), pg), pj)

### 5.3.5 Support Relation Nested within an Attack Relation

When the object attacked by proposition *po* is the support relation *S(pi, pj)* itself—rather than proposition *pi* or *pj*—the support relation shall be embedded as a whole within the attack relation.
**Representation:** A(po, S(pi, pj))
This structure indicates that *po* negates the supporting force of *pi* for *pj*, rather than directly negating *pi* or *pj*.
Multi-level nested structures shall be expressed step by step in accordance with the basic forms described above.
If the object of one relation is another relation, the latter shall be preserved in its entirety as a nested unit and then incorporated into the higher-level relational structure.
When expressing such structures, the hierarchy of parentheses shall strictly correspond to the hierarchy of logical structure. Cross-level merging or omission of intermediate structures is not permitted.

# 6. Graphical Representation Rules for Legal Argument Structures

## 6.1 Diagram Rules for Proposition Nodes

Each proposition corresponds to a node in the diagram. A proposition node is represented by a framed proposition identifier (e.g., p1, p2, …, pn). The proposition node itself does not convey information about the type of the proposition. Information about proposition types is recorded only in the annotation table or database and is not displayed in the diagram.

## 6.2 Graphical Representation of the Five Basic Structures

When no further relations occur between relation nodes (i.e., in a first-order structure), five types of relations—support relation, attack relation, joint relation, match relation, and identity relation—constitute the five basic structural patterns.

Except for the identity relation, each relation type is represented by a relation node. Proposition nodes are depicted as rectangles, while relation nodes are depicted as circles. Different fill styles distinguish relation types:

- Solid circles represent support relations;
- Hollow circles represent attack relations;
- Circles containing a "+" sign represent joint or match relations.

Table 8 below presents the specific graphical representations of these five basic structures.

**Table 8. Diagram Methods for the Five Basic Structures**

| Relation Type | Diagram Method | Example |
|---|---|---|
| **Support** | Represented by a directed edge with a solid circular node. The solid circle denotes the support relation node, and the arrow direction is: supporting element → supported element. | (p1 supports p2) |
| **Attack** | Represented by a directed edge with a hollow circular node. The hollow circle denotes the attack relation node, and the arrow direction is: attacking element → target of the attack. | (p1 attacks p2) |
| **Joint** | Represented by undirected links connected to a circle containing a "+" sign. The "+" circle denotes the joint relation node, indicating a conjunctive structure among multiple propositions. | (p1, p2, p3 joint) |
| **Match** | Represented by a directed edge with a circle containing a "+" sign. The "+" circle denotes the match relation node, and the arrow direction is: matching element → matched element. | (p1 matches p2) |
| **Identity** | Represented by a "/" symbol inside a rectangular node, indicating that the propositions are semantically identical. | p1/p2 (p1 and p2 are identical) |

## 6.3 Nested Diagramming Methods

### 6.3.1 Joint Relation Embedded within a Match Relation

When several propositions first form a **joint relation**, and this combined whole then participates as a single unit in a **match relation**, the diagram should adopt the method of ***Joint Relation Embedded within a Match Relation***.

**Diagramming Procedure:**

- First, use a circular node containing a "+" sign to represent the conjunctive (joint) structure.
- Then, draw a directed arrow from this "+" node to the target proposition to indicate the match relation.
- The "+" node simultaneously functions as the joint relation node, and the match relation node. In other words, no separate nodes are created for the two relations; the same "+" node carries both functions.

**Example 1**

Let: p3 be a **general proposition**, p1 and p2 be **particular propositions**. Propositions p1 and p2 form a joint relation, and this combined whole forms a match relation with p3.

**Formal expression:** M(J(p1, p2), p3)

**Diagram:**

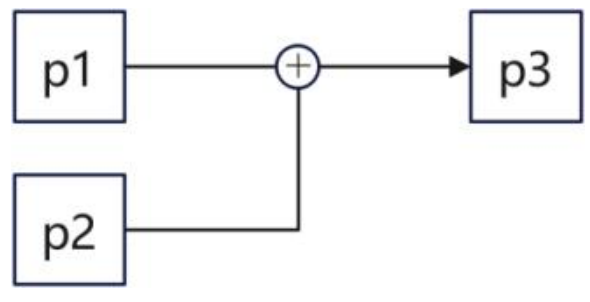

### 6.3.2 Match Relation Embedded within a Joint Relation

When several match relations as a whole further form a conjunctive structure, the diagram shall adopt the method of ***Match Relation Embedded within a Joint Relation***.

**Diagramming Procedure:**

- First, construct each match relation structure independently. Treat each match relation node as an independent relational unit.
- Introduce a new circular node containing a “+” sign to connect these match relation nodes into a joint relation. This “+” node represents the conjunctive relation among multiple matching results.

**Example 2**

Let: Particular proposition p1 and general proposition p2 form a match relation M1; Particular proposition p3 and general proposition p4 form a match relation M2; M1 and M2 then form a joint relation.

**Formal expression:** J(M1, M2), or J(M(p1, p2), M(p3, p4))

**Diagram:**

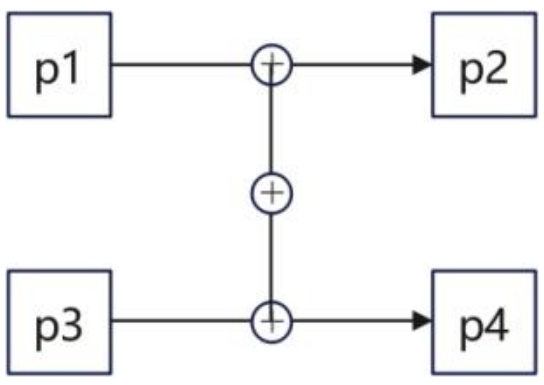

### 6.3.3 Joint Relation Embedded within a Support Relation

When several propositions must jointly hold in order to support a target proposition, the diagram shall adopt the method of ***Joint Relation Embedded within a Support Relation***.

**Diagramming Procedure:**

- First, use a circular node containing a “+” sign to represent the joint structure among several propositions. Treat this “+” node as a unified supporting unit.
- From this joint node, introduce a support relation node (a solid circle ●).
- The support relation arrow points to the supported proposition.

**Example 3**

Let the conjunction of propositions p1 and p2 support p3.

**Formal expression:** S(J(p1, p2), p3)

**Diagram:**

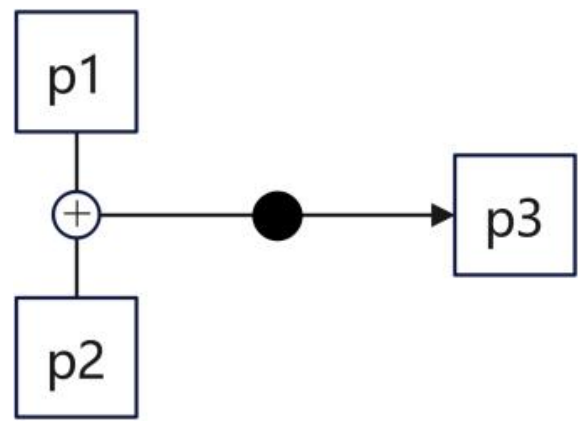

**Example 4**

Let the conjunction of propositions p1, p2, and p3 support p4.

**Formal expression:** S(J(p1, p2, p3), p4)

**Diagram:**

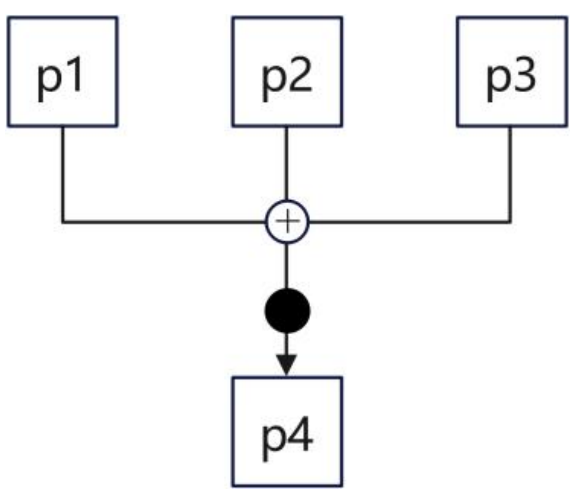

### 6.3.4 Match Relation Embedded within a Support Relation

When an entire match-relation structure serves as a supporting unit that further supports another proposition, the diagram shall adopt the method of ***Match Relation Embedded within a Support Relation***.

**Diagramming Procedure:**

- First, construct the match relation structure: the particular proposition points to the general judgment through a match relation node (a "+" circle).
- Then, treat the matched general judgment node as the starting point of the support relation (solid circle ●). The support relation arrow points to the supported proposition.

It should be noted that although, in the diagram, the support relation appears to originate from the general judgment node, the argumentative basis lies in the fact that the norm has already been satisfied in the particular case through the match relation. The diagram represents a structurally simplified expression and does not alter the hierarchical logic.

**Example 5**

Let particular proposition p1 match general proposition p2, and the conjunction of p1 and p2 support p3.

**Formal expression:** S(M(p1, p2), p3)

**Diagram:**

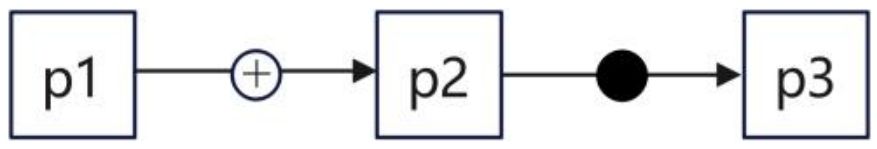

### 6.3.5 Support Relation Embedded within an Attack Relation

When a proposition does not directly attack another proposition, but instead attacks the **support relation** between two propositions, the diagram shall adopt the method of ***Support Relation Embedded within an Attack Relation***.

**Diagramming Procedure:**

- First, construct the support relation structure.
- Then, direct the attack-relation arrow toward the support relation node (●). The attack relation uses a hollow circle node (○), with the arrow direction: attacking proposition → attacked support-relation node.

**Example 6**

Let proposition p1 support proposition p2, and proposition p3 attack the support relation between p1 and p2.

**Formal expression:** **A**(p3, S(p1, p2))

**Diagram:**

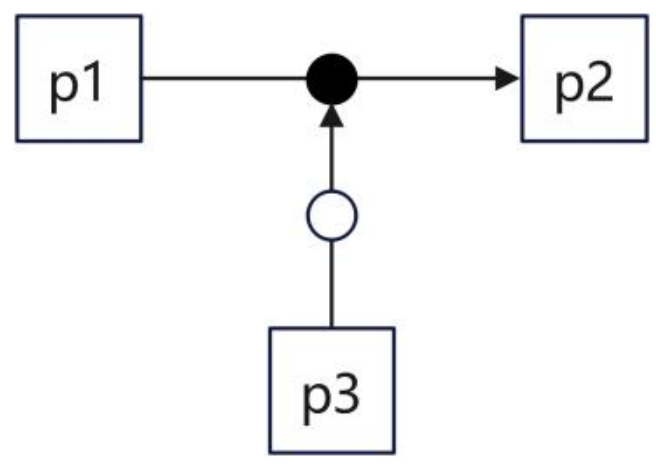

### 6.3.6 Multi-Level Nesting

When more than two levels of relational nesting coexist within an argumentative structure, the structure shall be constructed layer by layer from the innermost relation outward, maintaining the independence of each relation node at every level.

**Example 7**

Let particular propositions p1, p2, and p3 match general proposition p4, and the conjunction of p1, p2, p3, and p4 support proposition p5.

**Formal expression:** S(M(J(p1, p2, p3), p4), p5)

**Diagram:**

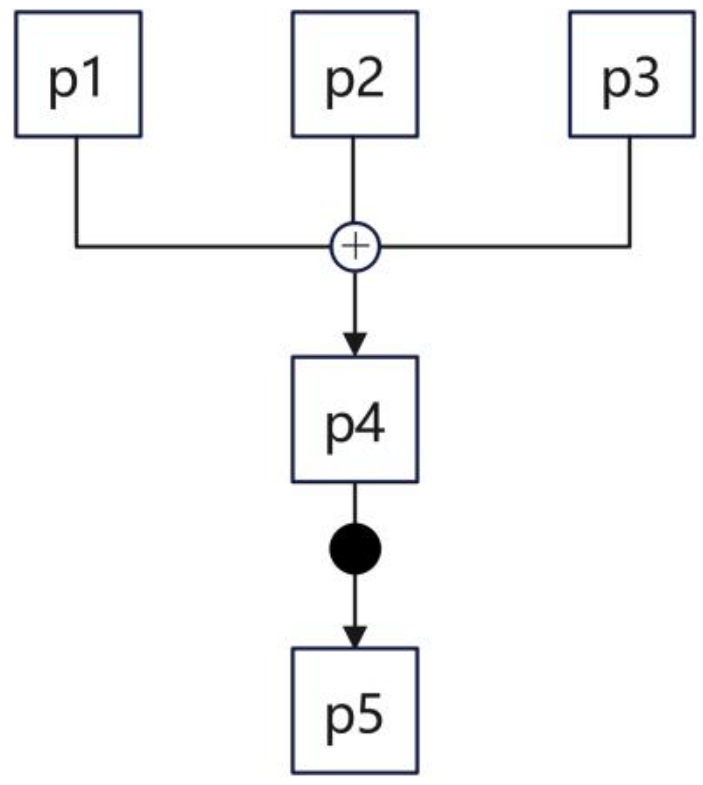

## 7. Example of Legal Argumentation Annotation: Full Annotation Process for Document I

## 7.1 Segmentation of the Reasoning Section of Judicial Decisions

According to the Guideline, the reasoning section of the judicial decision is segmented as follows:

> A contract formed in accordance with the law shall take effect upon its formation. The parties shall fully perform their respective obligations in accordance with their agreement. Where one party fails to pay the price, remuneration, rent, interest, or fails to perform any other monetary obligation, the other party may request payment. In the present case, the plaintiff and the defendant are in a labor service contract relationship. The plaintiff has provided labor services as agreed, and the defendant shall pay labor remuneration in accordance with the agreement between the parties. Based on the IOU and the WeChat transfer records submitted by the plaintiff, it can be determined that the defendant still owes the plaintiff RMB 11,000 in labor remuneration, and the defendant shall pay this RMB 11,000 to the plaintiff. The plaintiff further claims an additional RMB 600 in labor remuneration; however, the evidence provided is insufficient to prove this claim. Therefore, the court does not support the plaintiff's request that the defendant pay the additional RMB 600 in labor remuneration.

## 7.2 Element Segmentation, Numbering, and Type Annotation

**Table 9. Element Segmentation, Numbering, and Type Annotation for Document I**

| Element No. | Element Content | Type |
|---|---|---|
| P1 | A contract formed in accordance with the law shall take effect upon its formation. | GM-L |
| P2 | The parties shall fully perform their respective obligations in accordance with their agreement. | GM-L |
| P3 | Where one party fails to pay the price, remuneration, rent, interest, or other monetary debts, the other party may request payment. | GM-L |
| P4 | In the present case, the plaintiff and the defendant are in a labor service contract relationship. | SM |
| P5 | The plaintiff has provided labor services as agreed. | SF |
| P6 | and the defendant shall pay labor remuneration in accordance with the agreement between the parties. | SM |
| P7 | Based on the IOU and the WeChat transfer records submitted by the plaintiff, | SF |
| P8 | it can be determined that the defendant still owes the plaintiff RMB 11,000 in labor remuneration, | SF |
| P9 | and the defendant shall pay this RMB 11,000 to the plaintiff. | SM |
| P10 | The plaintiff further claims an additional RMB 600 in labor remuneration; | SF |
| P11 | however, the evidence provided is insufficient to prove this claim. | SF |
| P12 | that the defendant pay the additional RMB 600 in labor remuneration. | SM |
| P13 | Therefore, the court does not support the plaintiff's request | SM |

**Notes:**

(1) *In P4, the "labor service contract relationship" constitutes a legal relationship; therefore, P4 is labeled as SM (Particular Normative Judgment).*

(2) *Deontic expressions such as "shall" and "should" are typical indicators of normative judgments; therefore, P6 and P9 are labeled as SM.*

## 7.3 Relation Annotation

**Table 10. Relation Annotation for Document I**

| Relation Type | Relation Content |
|---|---|
| R1 | J(P4, P5); |
| R2 | M(J(P4, P5), P2) |
| R3 | S(M(J(P4, P5), P2), P6) |
| R4 | S(P7, P8) |
| R5 | J(P8, P6) |
| R6 | M(J(P8, P6), P3) |
| R7 | S(M(J(P8, P6), P3), P9) |

| R8 | S(P10, P12) |
|---|---|
| R9 | A(P11, P10) |
| R10 | S(P11, P13) |

## 7.4 Argument Diagram

This example contains two arguments, which are diagrammed separately.

**Diagram 1:**

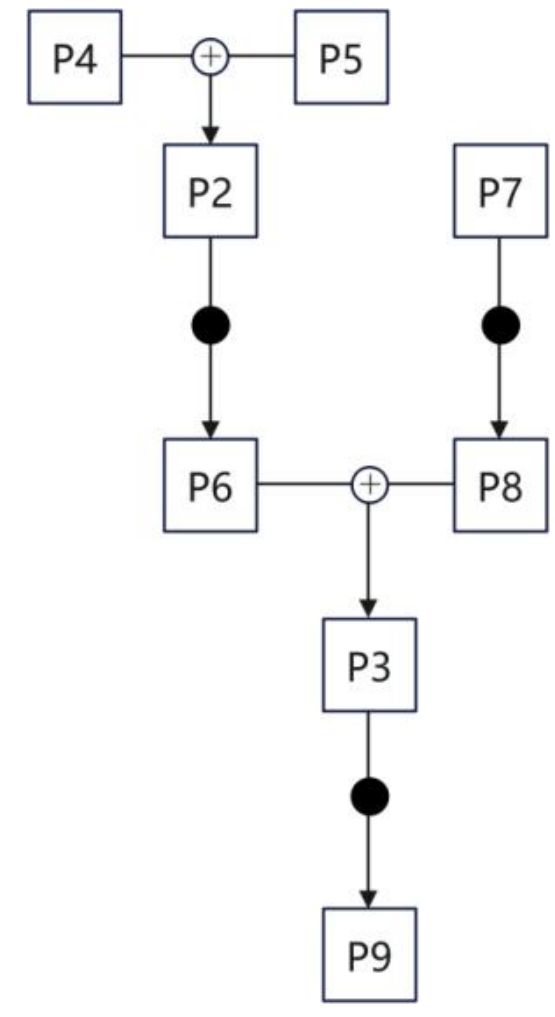

**Diagram 2:**

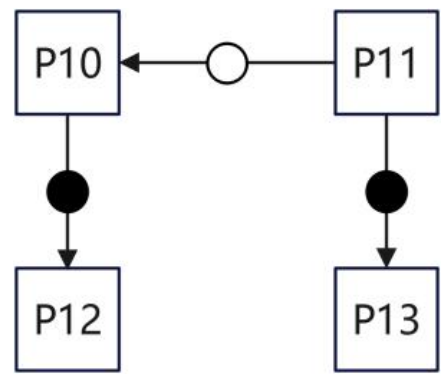

# 8. Annotation Workflow and Consistency Control

## 8.1 General Principles

The annotation work is guided by the fundamental principles of **operability, reproducibility, and verifiability**, and centers on the stable identification and consistent representation of legal argumentation structures. The workflow emphasizes:

- **Conceptual primacy:** All annotation labels must have clearly defined meanings grounded in legal argumentation theory. Annotation shall not rely merely on linguistic intuition.
- **Process constraints:** A staged and role-differentiated workflow design is adopted to reduce the impact of individual interpretative differences on annotation outcomes.
- **Built-in consistency control:** Consistency testing is embedded throughout the annotation process, rather than treated solely as a post hoc statistical measure.

## 8.2 Annotation Workflow Design

The annotation workflow follows a four-stage structure: **Training — Pilot Annotation — Formal Annotation — Review and Revision.**

### 8.2.1 Annotator Training and Guideline Study

Prior to formal annotation, all annotators receive standardized training, including:

- Definitions and distinction criteria for each type of annotation object;
- A clause-by-clause explanation of the annotation Guideline, supplemented by positive and negative examples (where available);
- Common ambiguity scenarios and corresponding resolution principles.

After the training, annotators are required to independently read the annotation Guideline and complete a confirmation of understanding.

### 8.2.2 Pilot Annotation and Guideline Calibration

A selected set of judicial documents is used as pilot material, and multiple annotators independently complete annotations. The purpose of the pilot stage is not to generate usable data, but to:

- Test the operational feasibility of the annotation Guideline;
- Identify labels with high disagreement rates and highly uncertain cases;
- Detect rules that require further refinement or consolidation.

Pilot results are discussed collectively, and revision proposals are formulated. The Guideline is calibrated one or more times until disagreements are reduced to an acceptable range.

### 8.2.3 Formal Annotation and Dual Independent Annotation

During the formal annotation stage, a **dual independent annotation mechanism** is adopted:

- Each annotated text must be independently annotated by at least two annotators who are unaware of each other's results;
- Annotators must strictly follow the finalized version of the annotation Guideline and may not introduce personal rules *ad hoc*;
- Annotation results are stored in structured form to facilitate subsequent alignment and comparison.

### 8.2.4 Conflict Resolution and Expert Review

Dual-annotation results are compared automatically or semi-automatically to identify inconsistencies, which are categorized as follows:

- Inconsistent label selection;
- Inconsistent annotation boundaries;
- Inconsistent relational direction or target assignment.

For routine disagreements, annotators resolve differences through discussion and recorded justification. For disagreements involving core conceptual interpretations, the project lead or a domain expert is consulted for adjudication. Such cases are simultaneously incorporated into a *Difficult Issues Documentation* file.

## 8.3 Consistency Control Mechanisms

### 8.3.1 Consistency Metrics and Monitoring

During the annotation process, inter-annotator agreement metrics are calculated periodically to monitor the degree of consistency among annotators.

### 8.3.2 Continuous Feedback and Iterative Revision

Consistency results and representative disagreement cases are periodically fed back to annotators. Small-group discussions are conducted to strengthen shared understanding. Without compromising the comparability of existing data, necessary supplementary clarifications may be added to the annotation Guideline.

### 8.3.3 Version Control and Traceability

Versioned management is applied to the annotation Guideline, annotation data, and conflict adjudication records to ensure:

- Any annotation result can be traced back to the corresponding version of the Guideline;
- Major rule changes are associated with clear time points and scopes of applicability;
- Reliable foundations are provided for subsequent model training, evaluation, and reproducibility experiments.